\documentclass[cameraready]{Interspeech}

\usepackage{xcolor}
\usepackage{tikz}
\definecolor{tgfillg}{HTML}{D4EDDA}
\definecolor{tgborderg}{HTML}{28A745}

\definecolor{btnborder}{HTML}{DEE2E6}
\definecolor{btnblue}{HTML}{A6CEE3}
\definecolor{btnorange}{HTML}{FDBF6F}

\DeclareRobustCommand{\redcircle}[1]{%
  \begin{tikzpicture}[baseline=-0.7ex]
    \node[
      shape=circle, 
      fill=red, 
      inner sep=0.5pt, 
      minimum size=1.8ex
    ] (char) {\fontsize{6.5}{7}\selectfont\textcolor{black}{\textup{#1}}};
  \end{tikzpicture}%
}

\DeclareRobustCommand{\tgcircleg}{%
  \begin{tikzpicture}[baseline=-0.7ex]
    \node[
      shape=circle, 
      draw=tgborderg, 
      fill=tgfillg,
      inner sep=0pt, 
      line width=0.4pt,
      minimum size=1.8ex
    ] (char) {%
      \fontsize{5}{6}\selectfont\sffamily\bfseries%
      \kern0.5pt\raisebox{0.3pt}{\textup T}\kern-0.5pt\raisebox{-0.3pt}{\textup G}\kern0.5pt%
    };
  \end{tikzpicture}%
}

\usepackage{cleveref}

\title{Speech Playground: An Interactive Tool for Speech Analysis and Comparison}

\author[affiliation={1}, orcid=0009-0009-4868-2868]{Stephen}{McIntosh}
\author[affiliation={1}, orcid=0000-0002-6265-9674]{Daisuke}{Saito}
\author[affiliation={1}, orcid=0000-0002-8778-9555]{Nobuaki}{Minematsu} 

\address{
    $^1$ The University of Tokyo, Japan
}

\email{\{smcintosh,mine\}@gavo.t.u-tokyo.ac.jp}

\usepackage{comment}

\begin{document}

\maketitle

\begin{abstract}
    This paper presents Speech Playground, an interactive speech visualization and comparison tool. While existing tools such as Praat are excellent, it can be cumbersome to integrate them with modern deep learning representations and use them for comparison. Speech Playground addresses this by combining a Python backend with a web-based frontend for interactive exploration of multiple feature types, including continuous, discrete, and variable-length representations. It includes TextGrid and forced alignment support together with configurable distance and alignment settings for visual and auditory comparison. Speech Playground is intended for use in speech research, representation validation, and computer-aided pronunciation training (CAPT)-oriented experimentation.
\end{abstract}

\section{Introduction}
Interactive tools for speech analysis such as Praat are widely used in speech research and are also useful for speech feedback in CAPT settings. 
However, recent deep-learning-based speech research has produced many different representations such as self-supervised or articulatory features. Comparing these requires Python-based encoders, alignment code, and ad-hoc visualization scripts, which is cumbersome.

\begin{figure}[t]
    \centering
    \includegraphics[width=\linewidth]{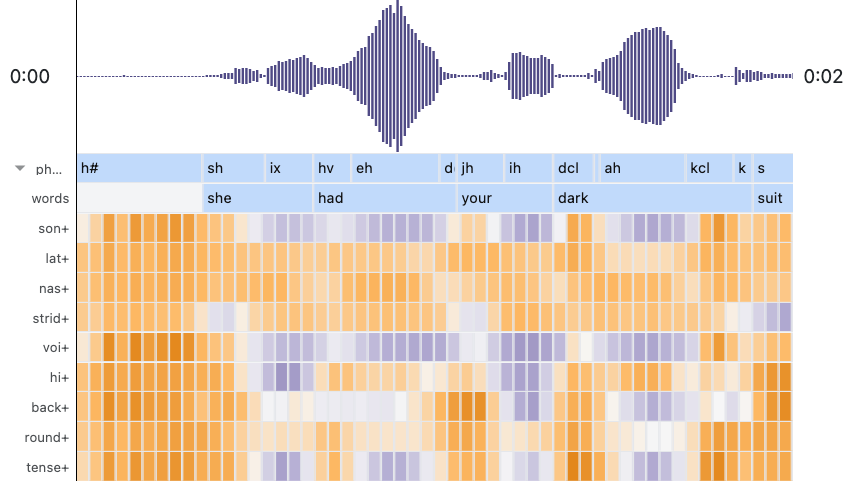}
    \caption{\textbf{Sample viewer} with TextGrid annotation and phonological vector tiers~\cite{choi2026bdt+p}. \textcolor{violet}{Positive} and \textcolor{orange}{negative} activations are shaded \textcolor{violet}{purple} and \textcolor{orange}{orange}, respectively.}
    \label{fig:analysis}
\end{figure}

In this paper, we present Speech Playground\footnote{\url{https://github.com/stephenmac7/speech-playground}}, an extensible interactive tool for visualization of speech features and utterance comparison. Speech Playground has two modes: \textit{Analysis} for single-track visualization (\Cref{fig:analysis}) and \textit{Diff} (\Cref{fig:ui}) for utterance comparison.

Speech Playground provides a single interactive environment in which users can compare speech encoders, continuous, discrete, and variable-length representations, and alternative distance and alignment settings on the same utterance pair.
We envision the following use cases: (1) \textbf{Speech research} using speech features unavailable in other tools and Diff mode to explain variation in speech with respect to a reference; (2) \textbf{Representation validation} by checking whether a representation captures a specific contrast or behaves consistently with the audio; and (3) \textbf{CAPT-oriented experimentation} using Diff mode to show where and how model speech and learner speech are different.

\section{Overview}

\subsection{Architecture}
Speech Playground comprises three components:

The \textbf{frontend} is a SvelteKit application that provides
two primary modes: \emph{Analysis}, for examining a single
utterance, and \emph{Diff}, for aligning and comparing two
utterances. WaveSurfer.js is used for waveform visualization.
IndexedDB is used to manage and persist uploaded recordings and metadata
such as transcription and TextGrid files.

The \textbf{backend} is a FastAPI (Python) server exposing
speech processing endpoints including encoding, segmentation, and alignment, lazily loading models on demand for fast startup and iteration.

The \textbf{speech-processing library} provides a uniform
interface over feature extractors, called \textit{encoders}.
Each encoder maps waveforms to a sequence of continuous frame- or segment-level representations.
Built-in encoders include SSL, articulatory, phonological-feature, and segmental representations, including SSL-derived variable-length representations such as ZeroSyl~\cite{visser2026zerosyl}.
Representations can optionally be transformed into discrete units or grouped into coarser variable-length segments.

The speech-processing library also includes functions for comparing utterances (used in \textit{Diff} mode).
It includes functions that compute similarity matrices between utterances and perform discrete or segment-based alignments, producing diffs with insertions, deletions, and substitutions. Users can switch between distance measures and alignment settings, including global and semi-global matching. For fixed-rate representations, Speech Playground defaults to dynamic time warping (DTW) via \texttt{dtw-python}\footnote{\url{https://dynamictimewarping.github.io/python/}}; variable-length segmental representations and discrete tokenizations can be compared with alternative alignment methods.

\begin{figure*}[t]
    \centering
    \includegraphics[width=0.95\linewidth]{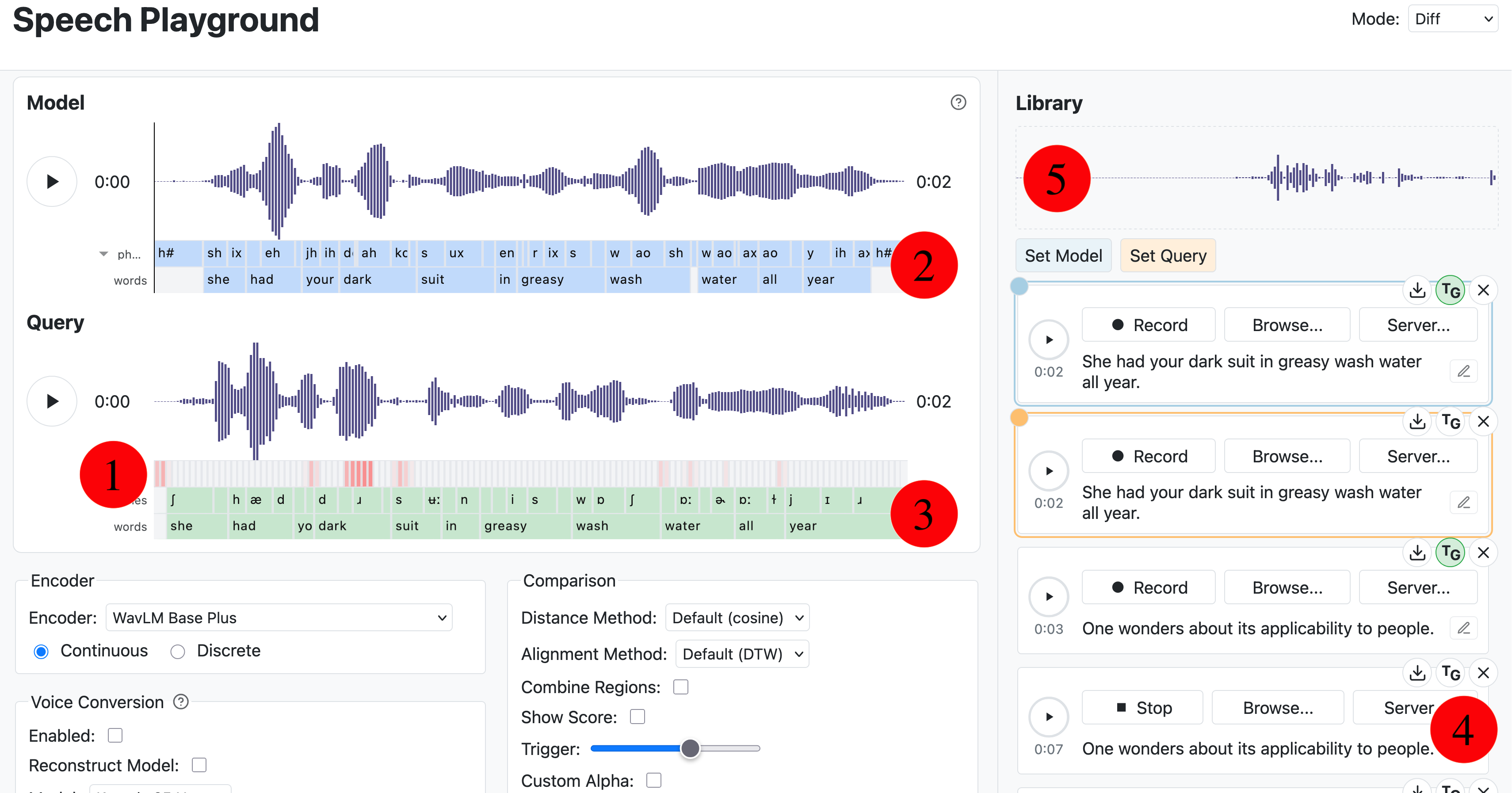}
    \vspace{-0.3em}
    \caption{\looseness=-1 \textbf{The full UI} in \textup{Diff} mode. The top tier in the \textbf{Query} \redcircle{1} shows the frame-wise DTW distance to the \textbf{Model} (higher distances in \textcolor{red}{red}). The blue tiers \redcircle{2} shown on the \textbf{Model} represent TextGrid tiers and are available for samples in the library with an attached TextGrid file (indicated by a green \tgcircleg{} button). The green tiers \redcircle{3} shown on the \textbf{Query} are a forced alignment using the optional MFA service.
    The user is currently recording new audio for the last ``One wonders about its applicability to people'' sample \redcircle{4} (recording progress shown at \redcircle{5}).}
    \label{fig:ui}
    \vspace{-1.8em}
\end{figure*}

\enlargethispage{2\baselineskip}
\subsection{Components}
\Cref{fig:ui} shows the full UI. It consists of a mode selector at the top right, a \textbf{library} in a sidebar on the right, and a main area on the left containing \textbf{sample viewer(s)} and \textbf{configuration}.

The \textbf{library} manages recordings and metadata (\textit{tracks}), including transcripts and TextGrid files.
Selected tracks are visualized in one or more \textbf{sample viewers}, which display the waveform together with interval tiers such as annotations, encoder-derived segments, or phonological features.
For tracks with transcripts, the viewer can also request a forced alignment\footnote{Forced alignment requires an additional backend server: \\\ \url{https://github.com/stephenmac7/mfa-service}}. The sample viewer is interactive: it can be zoomed and scrolled, and users can listen to segments of the audio by dragging over the waveform or intervals. In \textit{Diff} mode, holding \texttt{Shift} while selecting an area to play will play the corresponding audio in the \textit{other} track's sample viewer.

\subsection{Workflow}
\begin{figure}
    \centering
    \includegraphics[width=\linewidth]{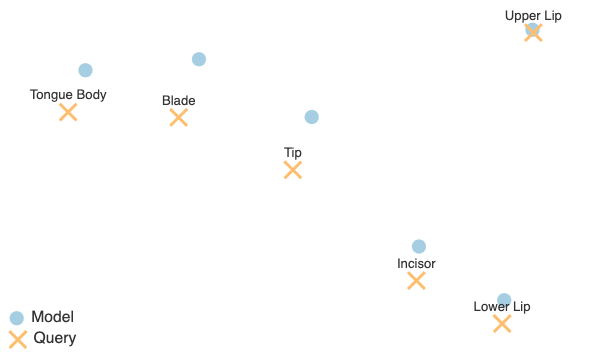}
    \caption{\textbf{Articulatory inversion} features~\cite{mcghee2025training} in Diff mode at a single frame. Animated when a sample plays.}
    \label{fig:art-inv}
    \vspace{-0.5em}
\end{figure}
Selected track(s) are encoded and compared (when in Diff mode) according to the selected configuration whenever either changes, allowing users to interactively switch encoders, discretization settings, distance measures, and alignment modes while exploring the results. Beyond locating mismatches, the aligned phonological and articulatory views can help users interpret how two utterances are different. For example, \Cref{fig:art-inv} shows articulatory inversion features aligned at a single frame, allowing direct inspection of estimated articulator differences between the two utterances.
\section{Conclusion}
Speech Playground is an interactive tool for analyzing and comparing speech. Its extensible design makes it easy to compare different speech representations and processing strategies within a single interface, making it useful for speech research, representation validation, and CAPT-oriented experimentation.

\section{Generative AI Use Disclosure}
LLMs were used for coding assistance and final proofreading.

\bibliographystyle{IEEEtran}
\bibliography{mybib}

\end{document}